\documentclass[sigconf]{acmart}

\AtBeginDocument{%
\providecommand\BibTeX{{%
\normalfont B\kern-0.5em{\scshape i\kern-0.25em b}\kern-0.8em\TeX}}}

\copyrightyear{2022} 
\acmYear{2022}
\setcopyright{acmlicensed}\acmConference[CIKM '22]{Proceedings of the 31st ACM International Conference on Information and Knowledge Management}{October 17--21, 2022}{Atlanta, GA, USA}
\acmBooktitle{Proceedings of the 31st ACM International Conference on Information and Knowledge Management (CIKM '22), October 17--21, 2022, Atlanta, GA, USA}
\acmPrice{15.00}
\acmDOI{10.1145/3511808.3557385}
\acmISBN{978-1-4503-9236-5/22/10}

\usepackage{amsmath}
\usepackage{multirow}
\usepackage{multicol}
\usepackage{pifont}
\usepackage{enumitem}
\usepackage{subcaption} 
\usepackage{pgfplots}
\usepackage[font=small, skip=8pt]{caption}
\usepackage[noend]{algpseudocode}
\usetikzlibrary{pgfplots.groupplots}
\pgfplotsset{width=4cm,compat=1.9}
\usepackage[ruled, linesnumbered]{algorithm2e}

\begin{document}

\title[Learn Basic Skills and Reuse: Modularized Adaptive Neural Architecture Search (MANAS)]{Learn Basic Skills and Reuse: \\Modularized Adaptive Neural Architecture Search (MANAS)}

\settopmatter{authorsperrow=4}

\author{Hanxiong Chen}
 \affiliation{
  \institution{Rutgers University}
  \institution{New Brunswick, NJ, US}
  \country{}
    }
 \email{hanxiong.chen@rutgers.edu}

\author{Yunqi Li}
 \affiliation{
  \institution{Rutgers University}
  \institution{New Brunswick, NJ, US}
  \country{}
    }
 \email{yunqi.li@rutgers.edu}

\author{He Zhu}
 \affiliation{
  \institution{Rutgers University}
  \institution{New Brunswick, NJ, US}
  \country{}
    }
 \email{hz375@cs.rutgers.edu}

\author{Yongfeng Zhang}
 \affiliation{
  \institution{Rutgers University}
  \institution{New Brunswick, NJ, US}
  \country{}
    }
 \email{yongfeng.zhang@rutgers.edu}

\renewcommand{\shortauthors}{Hanxiong Chen, Yunqi Li, He Zhu, \& Yongfeng Zhang}

\begin{abstract}
Human intelligence is able to first learn some basic skills for solving basic problems and then assemble such basic skills into complex skills for solving complex or new problems. For example, the basic skills ``dig hole,'' ``put tree,'' ``backfill'' and ``watering'' compose a complex skill ``plant a tree''. Besides, some basic skills can be reused for solving other problems. For example, the basic skill ``dig hole'' not only can be used for planting a tree, but also can be used for mining treasures, building a drain, or landfilling. The ability to learn basic skills and reuse them for various tasks is very important for humans because it helps to avoid learning too many skills for solving each individual task, and makes it possible to solve a compositional number of tasks by learning just a few number of basic skills, which saves a considerable amount of memory and computation in the human brain. We believe machine intelligence should also capture the ability of learning basic skills and reusing them by composing into complex skills. In computer science language, each basic skill is a ``module'', which is a reusable network of a concrete meaning and performs a specific basic operation. The modules are assembled into a bigger ``model'' for doing a more complex task. The assembling procedure is adaptive to the input or task, i.e., for a given task, the modules should be assembled into the best model for solving the task. As a result, different inputs or tasks could have different assembled models, which enables Auto-Assembling AI (AAAI).

In this work, we take recommender system as an example and propose Modularized Adaptive Neural Architecture Search (MANAS) to demonstrate the above idea. Neural Architecture Search (NAS) has shown its power in discovering superior neural architectures. However, existing NAS mostly focus on searching for a global architecture regardless of the specific input, i.e., the architecture is not adaptive to the input. In this work, we borrow the idea from modularized neural logic reasoning and consider three basic logical operation modules: AND, OR, NOT. Meanwhile, making recommendations for each user is considered as a task. MANAS automatically assembles the logical operation modules into a network architecture tailored for the given user. As a result, a personalized neural architecture is assembled for each user to make recommendations for the user, which means that the resulting neural architecture is adaptive to the model's input (i.e., the user's past behaviors). Experiments on different datasets show that the adaptive architecture assembled by MANAS outperforms static global architectures. Further experiments and empirical analysis provide insights to the effectiveness of MANAS. The code is open-source at \url{https://github.com/TalonCB/MANAS}.
\end{abstract}



\maketitle

\section{Introduction}

Neural Architecture Search (NAS) has emerged as a popular solution for alleviating the pain of designing neural network architectures. Its goal is to automatically discover the optimal deep neural networks based on data-driven learning so that practitioners are provided with ready-made deep neural architectures without expert knowledge, which reduces the burden on manual network design. NAS has become a prevailing research field in various applications such as computer vision \cite{zoph2016neural,pham2018efficient,liu2018darts}, natural language processing~\cite{pham2018efficient, chen2021techniques,liu2018darts} and recommender systems~\cite{yao2020efficient,song2020towards,li2022autolossgen}, which has been shown to have the ability of generating deep neural architectures that outperform some human-designed architectures and achieve state-of-the-art performance.

Existing NAS methods mainly focus on searching for a global neural architecture to fit all the data of a problem, which means that all the training data share one global neural network structure. However, the optimal network structure could be different for different data samples. As a result, we hope the network generation process can be adaptive to the input of the data sample. Under the context of recommender systems, each user is considered as a task (i.e., making recommendations for the user), and the input for this task is the user's historical behaviors (i.e., predicting the user's future behavior based on the histories). As a result, the generated network structure should be ``personalized'' so that the generated network is different for different users.

In this work, we propose a \textbf{M}odularized \textbf{A}daptive \textbf{N}eural \textbf{A}rchite- cture \textbf{S}earch (MANAS) framework to demonstrate the idea of learning basic skills as neural modules and then automatically assemble the modules into a model for solving different tasks. Most importantly, the module assembling process is adaptive to the input of the data sample and the task at hand, since the modules should be assembled in different ways to solve different problems. 
Following the idea of neural logic reasoning \cite{chen2021neural,shi2020neural,chen2022graph}, we consider three basic logical operation skills AND, OR and NOT as the basic modules, which are used to assemble complex neural architectures. The simple and meaningful design of the modules makes them a good fit for architecture search and realizing architecture-level personalization.

In Neural Collaborative Reasoning (NCR) \cite{chen2021neural}, these logical modules are organized into a modularized neural architecture according to the input based on a manually defined rule. Instead, our architecture search algorithm aims to automatically search for the optimal combination of the input variables together with the basic logical modules to generate input-adaptive architectures. Technically, we improve from the Efficient Neural Architecture Search (ENAS)~\cite{pham2018efficient} model by allowing the input of the data sample as the input for the neural architecture generation process, so that the input of the data sample can steer the generation process (i.e., input-adaptive). One can treat this as a sentence generation model by taking input from the data sample to generate an output sequence. The inputs are from the data sample while the output is the generated neural architecture. In this case, our MANAS model can generate data-specific architectures to provide an architecture-level personalization for users. We apply one-shot training as well as batch training strategies to speed up the training process. Such auto-architecture search mechanism eliminates the limitation of NCR that uses a manually designed global architecture. As a result, our framework can learn to generate more flexible and personalized neural architectures that are adaptive to different user inputs.

Our contributions can be summarized as follows.
\begin{itemize}
    \item We demonstrate the idea of learning some basic skills as neural modules and then reusing the modules by assembling them into different models for solving different tasks.
    \item We propose a modularized adaptive neural architecture search framework MANAS, which allows NAS model to generate adaptive input-dependent architectures for each input.
    \item We design a search space based on logical modules, which allows the neural architecture to be learned from data instead of using a manually crafted global architecture.
    \item We demonstrate the effectiveness of MANAS on different datasets compared with human-defined models. Analysis of the models generated by MANAS provides insightful guidance for model design. 
\end{itemize}


In the following, Section~\ref{sec:related} presents related work, Section~\ref{sec:preliminaries} reviews neural collaborative reasoning as preliminaries, Section~\ref{sec:model} presents our model, Section~\ref{sec:experiments} provides experiments, and Section~\ref{sec:conclusion} concludes the work with outlooks for future work.

\section{Related Work}
\label{sec:related}
\subsection{Personalized Recommendation}
The idea of personalized recommendation is to tailor the recommended results to a specific user or a set of users~\cite{ge2022survey,zhang2020explainable}. To achieve this goal, researchers designed multiple algorithms to do representation learning on user preferences. Conventional matrix factorization based approaches~\cite{salakhutdinov2008bayesian, koren2009matrix, rendle2009bpr} decompose the user-item rating matrix into two low-rank matrices for identifying latent features. Users' preferences are represented as low-dimensional vectors for calculating the similarity scores with candidate items. As neural network becomes a powerful approach in recommender systems, many research works explore the ways to incorporate deep neural network to capture personalized information. Early pioneer work designed a two-layer Restricted Boltzmann Machine (RBM) to model users' explicit ratings on items \cite{salakhutdinov2007restricted}. Later, deep learning based recommendation algorithms~\cite{xue2017deep, cheng2016wide, zhang2017joint, chen2021neural} were proposed to capture non-linear information in user-item interactions. More works have been proposed to apply deep neural networks to learn personalized user representations by capturing various information, such as sequential information~\cite{hidasi2016session, kang2018self, chen2018sequential, tang2018personalized, wang2021counterfactual}, image information~\cite{mcauley2015image,zhang2017joint,chen2019personalized, geng2022improving, chen2017personalized}, textual information~\cite{zhang2017joint,zhang2014explicit, li2021personalized, geng2022recommendation, xiong2021counterfactual, li2020generate, zhang2018towards}, attribute information \cite{xian2021ex3}, knowledge information \cite{geng2022path,fu2021hoops,xian2020cafe,fu2020fairness, xian2019reinforcement}, causal information \cite{xu2021causal,xu2021deconfounded,xu2022dynamic,li2022causal,li2021towards,tan2021counterfactual} and relational information \cite{fan2019graph, chen2022graph, tan2022learning, li2021efficient}. Researchers also consider reinforcement learning in recommendation tasks~\cite{ge2022towards, pei2019value, ge2022toward}. Some works also explored neural architecture search for recommendation and click-through-rate prediction \cite{yao2020efficient,song2020towards,li2022autolossgen}. Though the aforementioned models, including the architecture search-based models and plenty of works that cannot be enumerated, have different designs, their personalization is mainly reflected on representation-level, i.e., they employ a global model architecture for all users, and personalization is only captured by the user's personalized vector representation.

Recently, neural collaborative reasoning~\cite{chen2021neural, shi2020neural, chen2022graph, zhu2021faithfully} has been proposed to model recommendation tasks as logical reasoning problems. They take logically constrained neural modules to mimic the logical operations in a fuzzy logical space, and then these modules can be organized together with the input variables based on manually crafted rules to formalize the recommendation problem as a logical satisfiability task. Though these works have shown state-of-the-art performance, the architecture design still depends on human domain knowledge, which may require extra human efforts when transferring to different scenarios. Besides, these works still perform personalization on representation-level. Nevertheless, the design of modularized logical network inspires us to explore the possibility of implementing architecture-level personalization, so that the model can be more adaptive and generalizable to new areas without manually designed logical rules.

\vspace{-1ex}
\subsection{Neural Architecture Search}
Neural architecture search (NAS) refers to a type of method that can automatically search for superior neural network architectures for different tasks. The main components for designing NAS algorithms include the \textit{search space}, the \textit{search techniques} and the \textit{performance estimation strategy}~\cite{elsken2019neural}.

\textbf{Search Space}. An ideal search space should contain sufficient distinct architectures
while potentially encompassing superior human-crafted structures~\cite{song2020towards,zoph2016neural}. Existing work can be roughly classified into the macro architecture search space~\cite{real2017large, jin2019auto, baker2016designing, brock2018smash,cai2018efficient} and the micro cell search space~\cite{pham2018efficient, liu2018darts,real2019regularized,zoph2018learning,nayman2019xnas}. Macro search space design is to directly discover the entire neural network while micro search space focuses on learning cell structures and assembling a neural architecture by stacking many copies of the discovered cells. The former considers more diversified structures but is quite time-consuming for finding good architectures due to the huge search space. The latter reduces the computation cost but may not target good architectures due to the inflexible network structure.

\textbf{Search Techniques}. The dominant architecture search techniques include random search~\cite{bergstra2012random}, Bayesian optimization~\cite{bergstra2011algorithms}, evolutionary algorithms~\cite{real2017large}, reinforcement learning~\cite{zoph2016neural, zoph2018learning, pham2018efficient}, gradient-based optimization~\cite{liu2018darts}. In recent work~\cite{chen2019renas}, the authors shows the effectiveness of combining different types of search techniques to better balance exploitation and exploration.

\textbf{Performance estimation strategy}. When a model is searched, we need a strategy to estimate the quality of this model. Due to the expensive cost of fully training each sampled model, many performance estimation strategies are proposed, such as low-fidelity estimation~\cite{zoph2018learning,real2019regularized}, one-shot algorithm by weight sharing~\cite{pham2018efficient,liu2018darts}, network morphism~\cite{real2017large} and learning curve extrapolation~\cite{baker2017accelerating}.

\section{Preliminaries}
\label{sec:preliminaries}
Before formally introducing our MANAS framework, we first briefly review the concepts of neural logical modules in neural collaborative reasoning (NCR) \cite{chen2021neural,shi2020neural}, which is a neural-symbolic framework that integrates learning and reasoning. Neural logical modules use three independent multi-layer perceptron (MLP) to represent propositional logic operations AND, OR and NOT, respectively. To grant logic meanings to these modules, they conduct self-supervised learning by adding logical regularization over the corresponding neural logical modules to enable these modules to behave as logic operators. For example, the following logic regularizer is added over the AND module to make it satisfy the idempotence law in propositional logic, i.e., $x \wedge x = x$:
\begin{equation}
r=\frac{1}{|\mathcal{X}|}\sum_{\textbf{x}\in \mathcal{X}} 1-Sim(\text{AND}(\textbf{x}, \textbf{x}),\textbf{x})
\end{equation}
where $x$ is the input variable; $\mathcal{X}$ represents the input space; $Sim(\cdot, \cdot)$ is the similarity function to measure the distance of two variables, which is the cosine similarity function in \cite{chen2021neural}. The idea of other logical regularizers corresponding to other logical laws are similar and can be seen in \cite{chen2021neural, shi2020neural}. 

With these logical modules, the next step is to formalize the user-item interactions as logic expressions, so that the recommendation problem can be transformed into the problem of predicting the probability that a logical expression is true. NCR uses Horn clause to depict the recommendation scenario. Specifically, it predicts the next user-item interaction by taking the clues from the conjunction of a user's historical interactions. For example, if a user $u$ interacted with a set of items $\{v_1, v_2, v_3\}$ and we want to predict if this user would interact with item $\{v_4\}$ in the future, then the question can be translated into the following expression:
\begin{equation}
\label{eq:ncr_logic}
    \textbf{e}_{u, v_1} \wedge \textbf{e}_{u, v_2} \wedge \textbf{e}_{u, v_3} \rightarrow \textbf{e}_{u, v_4}
\end{equation}
where $\textbf{e}_{u, v_i}$ is the encoded predicate embedding which represents the event of user $u$ interacted with item $v_i$; "$\rightarrow$" is the material implication operator\footnote{Material implication ($\rightarrow$) can be represented by basic operations: $x\rightarrow y \Leftrightarrow \neg x \vee y$} in propositional logic. The expression Eq.~\eqref{eq:ncr_logic} can be equivalently converted to $\neg \textbf{e}_{u, v_1} \vee \neg \textbf{e}_{u, v_2} \vee \neg \textbf{e}_{u, v_3} \vee \textbf{e}_{u, v_4}$ by applying De Morgan's Law\footnote{De Morgan's Law: $\neg(x\vee y) \Leftrightarrow \neg x \wedge \neg y; \neg(x\wedge y) \Leftrightarrow \neg x \vee \neg y$}. Then the Horn clause can be identically transformed into a neural architecture using the logical neural modules. The output is also a vector which represents the true/false of the expression. The final scoring function is a similarity function such as cosine that can measure the similarity between this output and the true vector $\textbf{T}$. This true vector is an anchor vector that is fixed after initialization and will not be updated during the training process. It is used to define the logical constant True in the high dimensional vector space. We can then rank items based on these scores to generate ranking lists for users. In this work, we use similar notation $\textbf{e}_i$ to represent the predicate "item $v_i$ is interacted".

\section{MANAS Framework}
\label{sec:model}
We have briefly introduced the background of neural architecture search, neural logical modules and the challenges of implementing adaptive NAS. In this section, we will give details of our MANAS framework in terms of search space design, search algorithm and prediction/recommendation workflow.

\subsection{Search Space Design}
\begin{figure}
    \centering
    \includegraphics[scale=1.0]{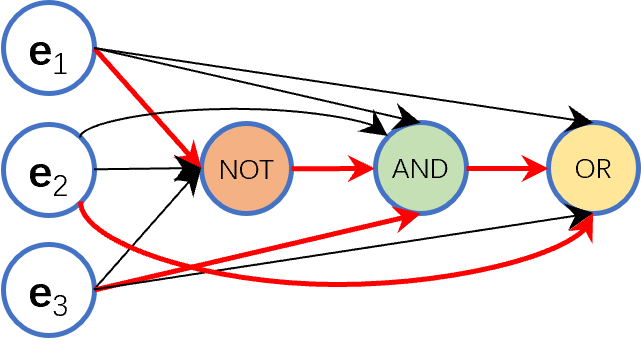}
    \caption{An illustration of an architecture in the designed search space. Red, green and yellow blocks are logical modules; white blocks are raw input variables. The arrows show some potential connections between blocks while the red lines show one possible valid architecture $\neg \textbf{e}_1 \wedge \textbf{e}_3 \vee \textbf{e}_2$.}
    \label{fig:example_1}
    \vspace{-10pt}
\end{figure}

In the NAS literature~\cite{liu2018darts, zoph2018learning}, it has been shown that determining the entire architecture of MLP can be extremely time-consuming. Thus, it is preferable to use a tailored but expressive search space that exploits domain-specific knowledge. As we mentioned before, this work can be treated as an extension of NCR, which is to apply NAS techniques and learn to generate modularized logical architectures for each input adaptively, thus realizing architecture-level personalization for the recommendation. In NCR, user historical events are always combined with a conjunction form. However, this form is not flexible and may not always fit the behavior patterns of each user.
In our work, we still use the Horn clause to predict a user's future behavior but replace the premise part (i.e. the left part of material implication) from a fixed conjunction architecture to a searched architecture that is adaptive to the user's interaction history.
Different from other NAS works which search for hyper-parameters and basic operators such as activation functions in neural architecture, we focus on searching for a valid combination of logical modules and input variables so that a personalized architecture can be searched and dynamically assembled according to the user's history to make recommendations for this user.

As shown in Fig.(\ref{fig:example_1}), for a given set of raw input variables $\textbf{e}_1, \textbf{e}_2, \textbf{e}_3$ and three logical modules AND, OR and NOT, the model should find a set of valid connections to combine all the raw inputs together with these logical modules to formulate a valid logical expression. The AND and OR modules are binary operators which take two inputs while NOT is a unary operator with only one input. 
To cover the cases where there are more than two variables in the expression, we allow the reuse of logical module blocks. 
We treat the connected modules as a set of directed acyclic graphs (DAGs) by following~\cite{pham2018efficient} and each logic module can take as input both raw variables and the intermediate hidden variables, i.e., the variables that are output from other blocks of higher topological order in the DAG.

One challenge is how to make sure the final searched architecture represents a valid logical expression. The most straightforward approach to define the search space is to allow the reuse of both raw input variables and the intermediate hidden variables. In this way, the model can potentially search for any possible combinations of the given input variables and logical modules. However, this would result in an infinite search space which is challenging for search algorithms. On the other hand, if we cannot guarantee to generate a valid logical architecture, the downstream child network cannot be built and trained. To solve the problem, we forbid the reuse of input variables (be it raw or hidden variables). This setting allows our design to be implemented with both correctness and efficiency.

In summary, our search space includes two types of blocks: neural logical module blocks and input variable blocks. Input variable blocks include the raw input variables and the intermediate hidden output variables. These variables are represented as fixed-length vectors and the logical modules as MLPs. The hyper-parameters of these blocks, such as the dimension of vectors, number of layers of MLPs, the activation function to be used, etc., are not considered to be part of the search space. We only search for the superior combination of these blocks to build valid logical architectures for recommendation. We allow the reuse of neural logical module blocks but do not allow the reuse of input variable blocks. For each searched AND or OR logical module block, the model needs to search two-variable blocks as their inputs. Then, for each searched input variable block, we need to determine if this variable needs to couple with a NOT logical module. Once all the variables, except for the final output, are consumed, the search process is done. Since we do not allow the reuse of input variable blocks, the total length of the searched logical expression should be $n$, where $n$ is the total number of raw input variables. The total number of layers for the entire architecture is $n-1$. To sample architecture for $n$ raw input variables, the total number of distinct architectures in the search space would be:
\begin{equation}
    \prod_{i=2}^n 4{i \choose 2} {2 \choose 1} = 4^{n-1}n!(n-1)!
\end{equation}
which is $O(n!)$ level. The designed search space contains plentiful distinct architectures and can cover most of the valid logic architectures, which allows the model to be adaptive to more recommendation patterns than NCR. 

\subsection{Search Strategy}
NAS usually requires full model training for performance evaluation which is extremely time-consuming. Motivated by the one-shot search algorithms in NAS~\cite{pham2018efficient, brock2018smash}, we allow parameter sharing among architectures so that we can apply one-shot algorithm on MANAS. Specifically, we follow ENAS \cite{pham2018efficient} by using reinforcement learning as the search technique. There are two groups of networks---controller network and child network.

\begin{figure}[t]
    \centering
    \includegraphics[scale=0.88]{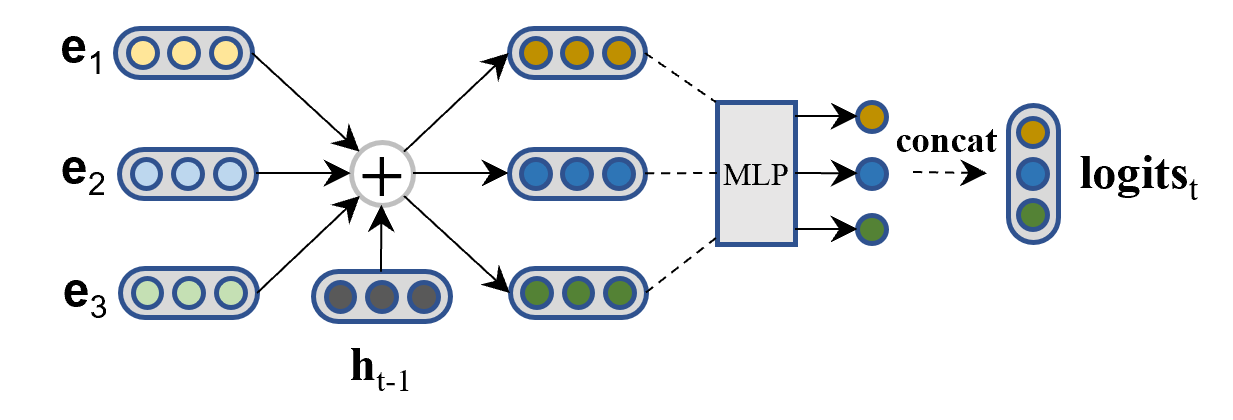}
    \vspace{-5pt}
    \caption{An illustration of the generation process of logits for input variable prediction at time $t$; $\textbf{e}_i$ represents predicate vector; $\textbf{h}_{t-1}$ is the hidden state from the $t$-$1$ step; ``+'' means element-wise addition.}
    \label{fig:input_logits}
    \vspace{-10pt}
\end{figure}

\subsubsection{\textbf{Controller Network}} Controller network is an LSTM~\cite{hochreiter1997long} which samples architecture modules via softmax classifiers. In ENAS, at each step, the controller takes the hidden state for the previous sampled module as input and makes the next prediction. The entire search process does not involve raw input variable vectors, which makes the architecture generation process independent from the input data. Different from ENAS, as illustrated in Fig.(\ref{fig:input_logits}), our controller network considers both the hidden states from the previous step and the input variable representations when making decisions. We assign a unique vector to each raw input variable and this is the reason why we treat each raw input variable as a block. For example, at time $t$, we have predicate vectors $\{\textbf{e}_1, \textbf{e}_2, \textbf{e}_3\}$ as the input candidates. We first merge the information from the hidden state and the input vectors by doing element-wise addition. Then these enriched vectors are sent to a mapping function whose output is a scalar. Each scalar can be treated as the logit of its corresponding input variable. We concatenate these logits to form a logits vector $\textbf{logits}_t$. This process can be represented as
\begin{equation}
\label{eq:logits}
\begin{split}
    l_i &= \mathbf{W}_1(\mathbf{e}_{i} + \mathbf{h}_{t-1}) + b_1 \\
    \mathbf{logits}_t &= \text{concat}(l_1, l_2, \ldots, l_i) \quad \forall i \in \mathcal{C}_t \\
\end{split}
\end{equation}
where $l_i$ is the logit for candidate $\textbf{e}_i$; $\textbf{W}_1\in \mathbb{R}^{n\times 1}$ is a matrix to map the input vector into a scalar; $b_1$ is the bias term; $\mathcal{C}_t$ represents the collection of candidates at current step. Then this $\textbf{logits}_t$ is sent to the Softmax function. Each dimension of the normalized logits vector represents the probability of a specific variable to be chosen as the input at the current time $t$. The prediction process of logical module is similar to the variable prediction process. The only difference is that we only use the hidden state from LSTM to create the logits vector for predicting the next module.

\begin{figure}[t]
    \centering
    \includegraphics[scale=0.65]{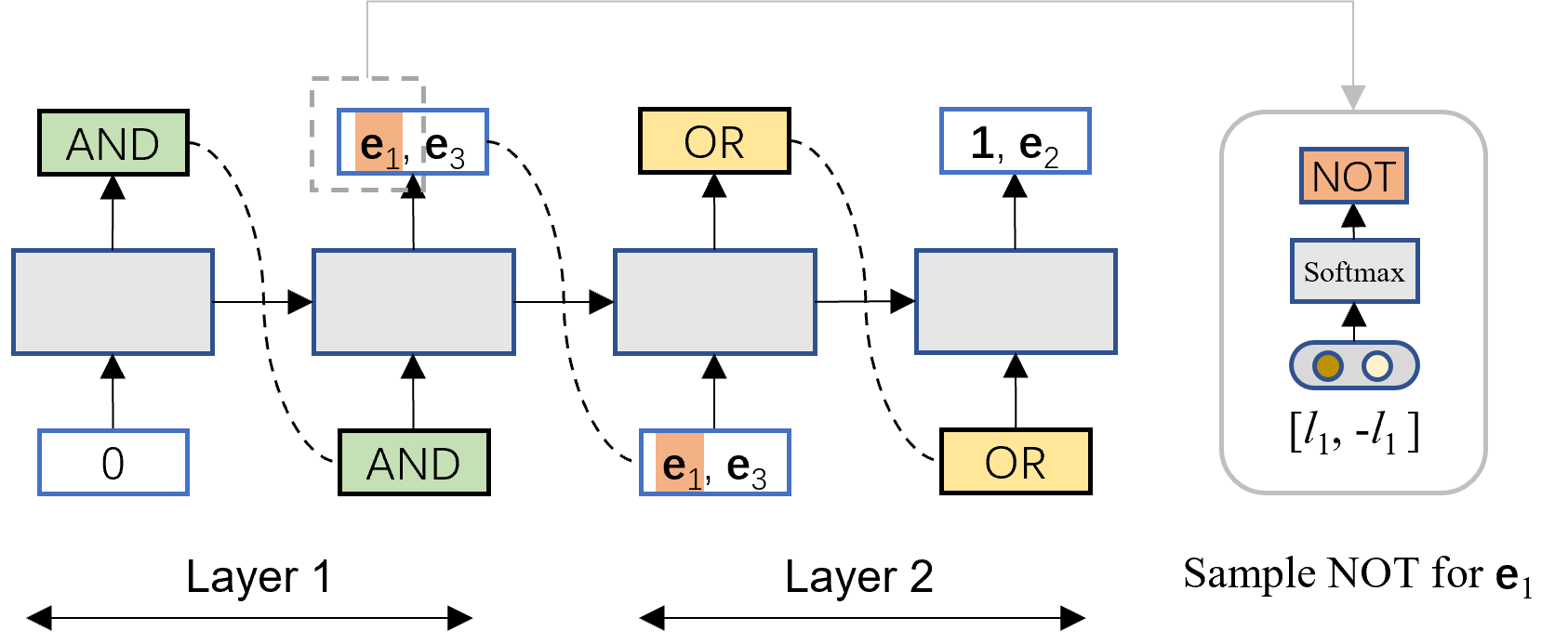}
    \caption{An illustration of the sampling process of logical architecture $(\neg \textbf{e}_1 \wedge \textbf{e}_3) \vee \textbf{e}_2$. $\textbf{e}_i$ represents predicate embedding; the first "0" means we use zero vector as the initial hidden state for LSTM; "$\textbf{1}$" in the last generation step represents the output hidden state from layer 1. The red background represents that the corresponding input vector is sent to the NOT module to get the negation form.}
    \label{fig:controller}
    \vspace{-10pt}
\end{figure}

Once we obtain the predicted input variables, the controller needs to decide if any of these inputs should be coupled with a NOT module to be represented as the negation form in logic expression. We take the logits of the selected input variables from the previous step to create a vector for NOT module prediction. For example, if we select $\textbf{e}_i$ as one of the input at step $t$, then we can calculate the logit $l_i$ by equation Eq.~(\ref{eq:logits}). Then the new logit vector of $\textbf{e}_i$ for NOT module prediction is $[-l_i, l_i]$, where the two dimensions represent do or do not couple with a NOT module, respectively. This is similar to what ENAS does for predicting skip connections. An example of the aforementioned sampling process is illustrated in Fig.(\ref{fig:controller}). 

\subsubsection{\textbf{Child Network}}
Child network is the searched network for the recommendation task. Since our approach is adaptive, for each input sequence of user interaction history, the controller network will first generate a data-specific architecture to organize the history sequence into a logic expression. Then we apply the Horn clause of the logical modules and input vectors to calculate the score for each candidate item. All the parameters of the child network are shared among all the sampled architectures. 

\subsection{MANAS Training and Deriving}
We have introduced the adaptive architecture search strategy and gave a brief illustration of the child network formalization process. To make it more clear and easy to understand, in this subsection, we will go through the training process of MANAS. 

In MANAS, the parameters of the controller network, denoted by $\theta$, and the shared parameters of the child network, denoted by $\omega$, are two groups of learnable parameters. These two networks are trained interleaved. First, we fix the controller's policy $\pi_\theta$ and train the child network on a whole pass of the training data, i.e. train one epoch. After that, we fix the parameters of the child network and train the controller's policy network parameters. To make the training process efficient, we apply one-shot searching algorithm since our child network parameters $\omega$ are shared. Additionally, we do not train child networks from scratch at each iteration. Instead, we update $\omega$ on top of the results of the last iteration. The training algorithm is given in Algorithm~\ref{alg:train}, and more details about MANAS training are as follows.

\subsubsection{\textbf{Child Network Training}}
In the recommendation task, we use a sequence of user interacted items as the input to predict future interactions. For each item $v_i$ in the item set $\mathcal{V}$, we assign a unique vector $\textbf{e}_i^\omega \in \textbf{E}_\omega$ to represent the predicate ``item $i$ is interacted,'' where $\textbf{E}_\omega$ represents the set of all the predicate vectors in the child network parameter space.

\begin{figure}[t]
    \centering
    \includegraphics[scale=0.8]{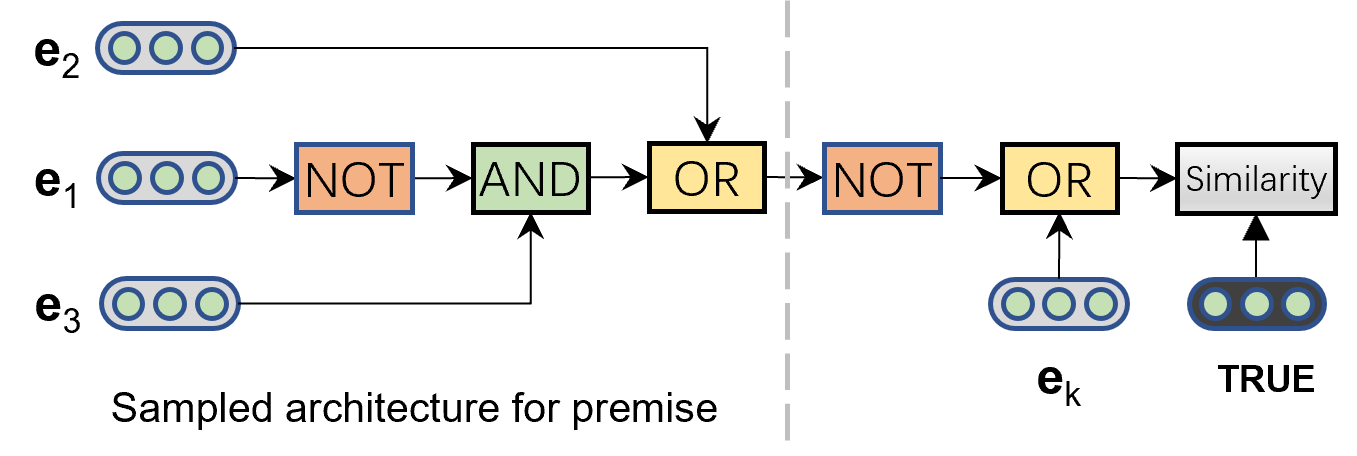}
    \caption{An illustration of a logical network for predicting the probability of item $k$ being interacted (represented by predicate $\textbf{e}_k$) with a sequence of input predicates $\{\textbf{e}_1, \textbf{e}_2, \textbf{e}_3\}$}.
    \label{fig:logic_network}
    \vspace{-20pt}
\end{figure}

Under the settings of neural collaborative reasoning \cite{chen2021neural}, the recommendation task is to answer the question ``$(?) \rightarrow v_k$,'' where $(?)$ is the premise logic expression that contains the input sequence and $v_k$ is a candidate item. Given a sequence $s=\{v_1, v_2, v_3\}$, the controller network will sample an architecture, e.g., $\neg \textbf{e}_1^\omega \wedge \textbf{e}_3^\omega \vee \textbf{e}_2^\omega$, from $\pi_\theta$ to convert the input into a logic expression with an empty embedding as the initial input. Then we replace ``(?)'' with this searched architecture and replace $v_k$ with its predicate vector $\textbf{e}_k^\omega$ to get the expression $\neg \textbf{e}_1^\omega \wedge \textbf{e}_3^\omega \vee \textbf{e}_2^\omega \rightarrow \textbf{e}_k^\omega$. Then we follow NCR to build and train the network by converting the expression into $\neg(\neg \textbf{e}_1^\omega \wedge \textbf{e}_3^\omega \vee \textbf{e}_2^\omega) \vee \textbf{e}_k^\omega$. The entire logical network of the given example is presented in Fig.(\ref{fig:logic_network}). We use the BPR-loss \cite{rendle2009bpr} by sampling one user non-interacted item as the negative sample for each ground-truth item to train the child network. We also use recommendation metrics such as hit ratio and normalized discounted cumulative gain (NDCG) as the performance evaluation measure. 

\begin{algorithm}[t]
\caption{MANAS Training Algorithm}
\label{alg:train}
\SetKwInOut{Input}{Input}
\SetKwInOut{Output}{Output}
\SetAlgoLined
\Input{Controller parameters $\theta$; child logical network shared parameters $\omega$; performance evaluator $\mathcal{E}$; training steps of controller network $K$;
training epochs $t$; training data $\mathcal{D}_{train}$; validation data $\mathcal{D}_{valid}$; optimizer OPTIM}
\BlankLine
Initialize controller parameters $\theta$ and child logical network parameters $\omega$\;

\For{$epoch \leftarrow 1$ \KwTo $t$} {
    sample architectures $\mathcal{M}_{train}$ from $\pi(s; \theta), \quad \forall s\in \mathcal{D}_{train}$\;
    train child logical network with $\mathcal{M}_{train}$ on $\mathcal{D}_{train}$\;
    \BlankLine
    \For{$k \leftarrow 1$ to $K$} {
        prepare $Batch$ from $\mathcal{D}_{valid}$\;
        \For{$b\in Batch$ } {
            sample architectures $\mathcal{M}_{b}$ from $\pi(s^\prime; \theta), \quad \forall s^\prime\in Batch$\;
            evaluate model with $\mathcal{E}$ to obtain reward $\mathcal{R}$\;
            update controller parameters $\theta$ with OPTIM\;
        }
    }
}
\end{algorithm}
\subsubsection{\textbf{Controller Network Training}}
In the controller network parameter space, we also assign a unique vector $\textbf{e}_i^\theta \in \textbf{E}_\theta$ for each item $v_i \in \mathcal{V}$. Though $\textbf{e}_i^\theta$ and $\textbf{e}_i^\omega$ are independent, they both represent the predicate ``item $v_i$ is interacted''. This can be a bridge to connect the controller network and the child network so that the generated architecture can be adaptive to the input sequences.
When training the controller network, we fix $\omega$ and update the policy network parameters $\theta$. Our goal is to maximize the expectation of reward $\mathbb{E}_{m\sim\pi(s; \theta)}[\mathcal{R}(m, \omega, s)]$, where $m$ represents a sampled child network; $s$ is the input sequence. The reward $\mathcal{R}(m, \omega, s)$ is computed on the validation set.
In our experiments, the reward function is the summation of hit ratio and NDCG on a minibatch of the validation set. We treat one whole pass of the validation set as one training step for the controller network and we train 50 steps per iteration. We employ the Adam optimizer~\cite{kingma2014adam}, for which the gradient is computed using REINFORCE, with a moving average baseline to reduce variance. 

\subsubsection{\textbf{Deriving Architecture}}
In ENAS, it only maintains the learned controller network parameters and discard the child network parameters when deriving new architectures. They sample a set of new architectures for validation and select the one with the best performance to train a new child network from scratch. Different from ENAS, we treat both the controller network and the child network as the components of MANAS. In testing time, when a new sequence $s^\prime$ is given, we directly sample a model $m_{s^\prime}$ from the trained policy $\pi(s^\prime; \theta)$. Then the child network $m_{s^\prime}$ is assembled based on the sampled architecture by using the trained shared logical modules and predicate vectors.

\section{Experiments}
\label{sec:experiments}
In this section, we evaluate MANAS as well as several baseline models for the top-$K$ ranking problem on a set of publicly available real-world recommendation datasets. We aim to answer the following research questions:
\begin{itemize}
    \item \textbf{RQ1}: Can MANAS search for architectures that outperform state-of-the-art human-crafted models for recommendation?
    \item \textbf{RQ2}: In MANAS, the searched architecture is adaptive to the input sequence. Is this adaptive nature important for improving recommendation quality?
    \item \textbf{RQ3}: The architecture sampling process in MANAS involves exploration since the modules are sampled according to logits probability rather than directly selecting the module of the largest probability (i.e., exploitation). Is exploration necessary for training MANAS?
    \item \textbf{RQ4}: What about the efficiency of MANAS?
    \item \textbf{RQ5}: 
    How does the recommendation performance and the training time cost change w.r.t the input sequence length?
\end{itemize}

\subsection{Experimental Settings}
\subsubsection{\textbf{Dataset}} In the experiments, we use the publicly available Amazon e-commerce dataset\footnote{https://jmcauley.ucsd.edu/data/amazon/}, which includes the user, item and rating information. We take \textbf{Beauty}, \textbf{Cell Phones and Accessories} and \textbf{Grocery and Gourmet Food} sub-categories for our experiments to evaluate the performance of our neural architecture search algorithm. Statistics of the datasets are shown in Table~\ref{tb:dataset}. 

\subsubsection{\textbf{Evaluation Protocol}}
We use leave-one-out strategy to split the dataset into train, validation and test, which is a widely used approach in the literature~\cite{chen2021neural,shi2020neural}. For each input sequence $s_u$ from a user $u$, we use the most recent interaction of each user for testing, the second recent item for validation, and the remaining items for training. As the searched neural architectures are used for ranking tasks, for efficiency consideration, we use \textit{real-plus-N} \cite{bellogin2011precision,said2014comparative} to calculate the measures in the validation and testing stage. More specifically, for each user-item pair in the validation and test set, we randomly sample 99 user non-interacted items, and we rank these 100 items for the ranking evaluation. 

As for the evaluation metrics, we select a recall-based metric Hit Ratio (Hit@$K$) and a rank-position based metric normalized discounted cumulative gain (NDCG@$K$) for evaluating the performance of recommendation. The reported results of all metrics are averaged over all users.

\begin{table}[t]
\caption{Statistics of the datasets in our experiments.}\label{tb:dataset}
\vspace{-5pt}
\begin{tabular}{ccccc}
\toprule
Dataset& \#Users &\#Items &\#Interaction & Density\\
\midrule
Beauty &22,363 &12,101 &198,502 &0.073\%\\ 
Cellphones &27,879 &10,429 &194,439 &0.067\%\\
Grocery &14,681 &8,713 &151,254 & 0.118\%\\
\bottomrule
\end{tabular}
\vspace{-10pt}
\end{table}

\subsubsection{\textbf{Hyper-parameters Settings}}
For our MANAS model, the length of the input sequence is 4, i.e., each user-item pair comes with 4 histories. In practice, one can set this number to any value but our experiments show that 4 history is good enough for our recommendation tasks. We will also explore the effect of different input lengths in Section \ref{sec:sequence_length}. For the baseline models, we allow sequence length up to 20. The embedding size is set to 64 for both the child network and the controller LSTM hidden vectors. We use Adam~\cite{kingma2014adam} to optimize model parameters. $\ell_2$ regularization is adopted to prevent child network from overfitting and the weight of $\ell_2$ is set to $10^{-5}$. The logic regularization weight is set to $10^{-5}$. The learning rate for the child network is 0.001 while the controller learning rate is 0.005.

\begin{table*}[t]
\caption{Results of recommendation performance on three datasets with metrics NDCG (N) and Hit Ratio (HR). We use underline (\underline{number}) to show the best result among the baselines. We use bold font to mark the best result of the whole column. We use one star (*) to indicate that the performance is significantly better than the best non-NAS based baselines, and use two stars (**) to indicate that the performance is significantly better than all baselines including NANAS. The significance is at 0.01 level based on paired $t$-test. Improvement$^1$ shows our model improvement over the best result (i.e., over \underline{number}), while improvement$^2$ shows our model improvement over NCR.}
\label{tb:results}
\setlength{\tabcolsep}{2pt}

\centering\begin{tabular}{lcccccccccccc}
\toprule
\multirow{2}{*}{} &\multicolumn{4}{c}{\textbf{Beauty}} & \multicolumn{4}{c}{\textbf{Cellphones}} & \multicolumn{4}{c}{\textbf{Grocery}} \\ 
\cmidrule(lr){2-5}
\cmidrule(lr){6-9}
\cmidrule(lr){10-13}
&N@5 &N@10 &HR@5 &HR@10 &N@5 &N@10 &HR@5 & HR@10 &N@5 &N@10 &HR@5 &HR@10 \\
\midrule
 GRU4Rec &0.1656 &0.2012 &0.2534 &0.3637 &0.1766 &0.2101 &0.2575 &0.3616 &0.2000 &0.2292 &0.2914 &0.3816\\
 NARM & 0.1900 & 0.2200 &0.2656 &0.3584 &0.2008 &0.2387 &0.2923 &0.4101 &0.2260 &0.2541 &0.3129 &0.3999\\
 Caser &0.2107 &0.2416 &0.2912 & 0.3874 &0.2054 &0.2412 &0.2937 & 0.4052 & 0.2313 &0.2603 & 0.3191 & 0.4088 \\
 SASRec & \underline{0.2599} & \underline{0.2861} & 0.3275 & 0.4091 & 0.2458 & 0.2767 & 0.3275 & 0.4230 & 0.2342 & 0.2658 & 0.3379 & 0.4361 \\
 BERT4Rec & 0.2283 & 0.2584 & 0.3102 & 0.4034 & \underline{0.2536} & 0.2908 & \underline{0.3503} & 0.4655 & 0.2420 & 0.2727 & 0.3356 & 0.4310 \\
\midrule
 NCR &0.1883 &0.2163 &0.2551 &0.3420 & 0.1819 & 0.2140 &0.2604 & 0.3602 & 0.2210 & 0.2490 &0.3140 & 0.4010\\ 
NANAS & 0.2519 & 0.2820 &\underline{0.3383} &\underline{0.4316} & 0.2534 & \underline{0.2915} & 0.3500 & \underline{0.4682} & \underline{0.2462} & \underline{0.2731} & \underline{0.3574} & \underline{0.4402}\\ 
\textbf{MANAS} &\textbf{0.2618} &\textbf{0.2933**} &\textbf{0.3532**} &\textbf{0.4507**} &\textbf{0.2785**} & \textbf{0.3149**} &\textbf{0.3846**} & \textbf{0.4971**} & \textbf{0.2609**} & \textbf{0.2882**} &\textbf{0.3598*} & \textbf{0.4444*}\\ 

\midrule
\midrule
Improvment$^1$ & 0.73\% & 2.52\% & 4.40\% & 4.43\% & 9.82\% & 8.03\% & 9.79\% & 6.17\% & 5.97\% & 5.53\% & 0.67\% & 0.95\%\\

Improvment$^2$ & 39.03\% & 35.60\% & 38.46\% & 31.78\% & 53.11\% & 47.15\% & 47.70\% & 38.01\% & 18.05\% & 15.74\% & 14.59\% & 10.82\%\\
 \bottomrule

\end{tabular}
\vspace{-5pt}
\end{table*}

\subsection{Baselines}
Since our model requires interaction sequence as input, we select the following representative sequential recommendation baselines to verify the effectiveness of our method.
\begin{itemize}
    \item \textbf{GRU4Rec} \cite{hidasi2016session}: This is a sequential/session-based recommendation model, which uses Recurrent Neural Network (RNN) to capture the sequential dependencies in users' historical interactions for recommendations.
    \item \textbf{NARM}~\cite{li2017NARM}: This model utilizes GRU and the attention mechanism to consider the importance of interactions . 
    \item \textbf{Caser}~\cite{tang2018personalized}: This is a convolutional neural network based sequential model, which learns sequential patterns using vertical and horizontal convolutional filters.
    \item \textbf{SASRec}~\cite{kang2018self}: This model uses transformer to capture the left-to-right context information from historical interactions.
    \item \textbf{BERT4Rec}~\cite{sun2019bert4rec}: This model utilizes a bi-directional self-attention module to capture context information in user behavior sequences from both left-to-right and right-to-left.
    \item \textbf{NCR}~\cite{chen2021neural}: This is a state-of-the-art neural logic reasoning based recommendation framework. It utilizes logic reasoning to model recommendation tasks. 
    \item \textbf{NANAS}: This is the non-adaptive version of our NAS model. It is used to replicate regular NAS which learns a global architecture for all inputs on a specific task. 
\end{itemize}

The baseline implementations are from an open-source recommendation toolkit \cite{wang2020make}. We select the best model based on its best performance on the validation set. We implement the models with PyTorch v1.8 and the models are trained on a single 2080Ti GPU.

\subsection{Recommendation Performance (RQ1)}
We report the overall recommendation ranking performance of all the models in Table~\ref{tb:results}. The results show that our MANAS consistently outperforms all the baseline models. From these results, we have the following observations.

(1) NCR outperforms GRU4Rec in most cases on all the datasets and achieve a comparable performance with NARM and Caser on \textit{Grocery} dataset. However, on \textit{Beauty} and \textit{Cellphones} datasets, Caser has a much better performance than NCR. This observation indicates that the manually designed architectures are lack of adaptation ability especially for neural logic reasoning. According to the findings in NCR, a correct form of the logic expression could affect the recommendation quality. However, designing a data-specific architecture is non-trivial and requires excessive human efforts, which corroborates the necessity of designing an data-adaptive neural architecture search algorithm.
    
(2) MANAS can consistently outperform all the baselines. By comparing MANAS and NCR, the significant improvement on all the metrics show that learning to generate architectures does help neural logic reasoning to be more adaptive to various inputs.

\begin{table*}[t]
\caption{Results of ranking performance in terms of average, min and max over 20 sampled architectures for each input sequence with MANAS. Non-sample model represents the performance of MANAS without exploration in the searching process. We mark $\textnormal{MANAS}_{AVG}$ in bold to show our model consistently outperforms non-sample model. STD is the standard deviation of the reported results on the 20 sampled architectures.}
\label{tb:non_sample}
\centering\begin{tabular}{lcccccccccccc}
\toprule
\multirow{2}{*}{} &\multicolumn{4}{c}{\textbf{Beauty}} & \multicolumn{4}{c}{\textbf{Cellphones}} & \multicolumn{4}{c}{\textbf{Grocery}} \\ 
\cmidrule(lr){2-5}
\cmidrule(lr){6-9}
\cmidrule(lr){10-13}
&N@5 &N@10 &HR@5 &HR@10 &N@5 &N@10 &HR@5 & HR@10 &N@5 &N@10 &HR@5 &HR@10 \\
\midrule
 NCR &0.1883 &0.2163 &0.2551 &0.3420 & 0.1819 & 0.2140 &0.2604 & 0.3602 & 0.2210 & 0.2490 &0.3140 & 0.4010\\ 
Non-Sample &0.2528 &0.2819 &0.3366 &0.4269 & 0.2627 & 0.2987 &0.3615 & 0.4734 & 0.2373 & 0.2674 &0.3456 & 0.4378\\ 
\midrule
$\textbf{MANAS}_{AVG}$ &\textbf{0.2618} &\textbf{0.2933} &\textbf{0.3532} &\textbf{0.4507} &\textbf{0.2785} & \textbf{0.3149} &\textbf{0.3846} & \textbf{0.4971} & \textbf{0.2609} & \textbf{0.2882} &\textbf{0.3598} & \textbf{0.4444}\\ 
$\textbf{MANAS}_{MIN}$ &0.2612 & 0.2923 &0.3523 &0.4485 &0.2775 & 0.3138 &0.3830 & 0.4951 & 0.2596 & 0.2869 &0.3584 & 0.4432\\
$\textbf{MANAS}_{MAX}$ &0.2638 & 0.2954 &0.3545 &0.4526 &0.2791 & 0.3155 &0.3859 & 0.4980 & 0.2621 & 0.2893 &0.3619 & 0.4462\\
\midrule
\midrule
STD & 0.0007 & 0.0008 & 0.0010 & 0.0013 & 0.0006 & 0.0005 & 0.0010 & 0.0008 & 0.0007 & 0.0008 & 0.0011 & 0.0013\\

 \bottomrule

\end{tabular}
\end{table*}
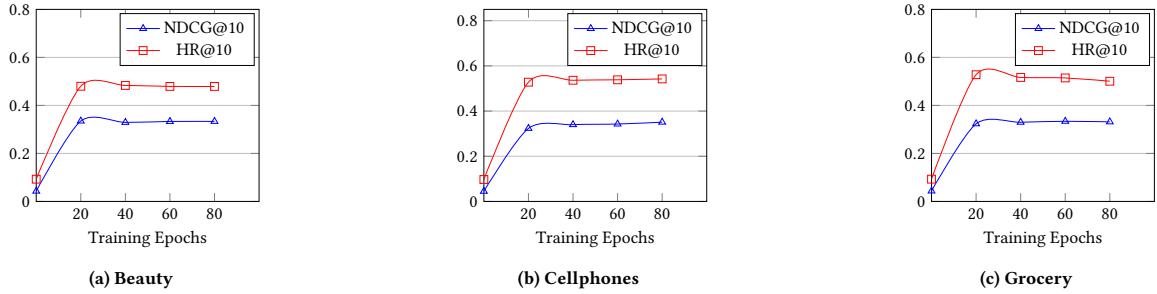
\begin{figure*}[t]
\begin{subfigure}[b]{0.33\textwidth}
    \centering
    \resizebox{0.6\linewidth}{!}
    {
        \begin{tikzpicture}
        \pgfplotsset{
            scale only axis,
            label style={font=\normalsize}
        }
            \begin{axis}[
                xlabel=Training Epochs,
                xmin=0, xmax=100,
                xtick={20, 40, 60, 80},
                xticklabels={20, 40, 60, 80},
                ymajorgrids=true,
                ymin=0, ymax=0.8
            ]
            
            \addplot[smooth,color=blue,mark=triangle]
                coordinates {
                    (0,0.0428)
                    (20,0.3348)
                    (40,0.3292)
                    (60,0.3332)
                    (80,0.3333)
                };
            \addlegendentry{NDCG@10}
            
            \addplot[smooth,mark=square,red] plot coordinates {
                (0,0.0927)
                (20,0.4798)
                (40,0.4832)
                (60,0.4791)
                (80,0.4790)
            };
            \addlegendentry{HR@10}

            \end{axis}
        \end{tikzpicture}
    }
    \caption{Beauty}   
    \label{fig:subfig_beauty}
\end{subfigure}
\begin{subfigure}[b]{0.33\textwidth}
    \centering
    \resizebox{0.6\linewidth}{!}
    {
        \begin{tikzpicture}
        \pgfplotsset{
            scale only axis,
            label style={font=\normalsize}
        }
            \begin{axis}[
                xlabel=Training Epochs,
                xmin=0, xmax=100,
                xtick={20, 40, 60, 80},
                xticklabels={20, 40, 60, 80},
                ymajorgrids=true,
                ymin=0, ymax=0.85,
                legend style={
                }
            ]
            
            \addplot[smooth,color=blue,mark=triangle]
                coordinates {
                    (0,0.0449)
                    (20,0.3233)
                    (40,0.3398)
                    (60,0.3426)
                    (80,0.3508)
                };
            \addlegendentry{NDCG@10}
            
            \addplot[smooth,mark=square,red] plot coordinates {
                (0,0.0971)
                (20,0.5278)
                (40,0.5364)
                (60,0.5384)
                (80,0.5423)
            };
            \addlegendentry{HR@10}
            
            \end{axis}
        \end{tikzpicture}
    }
    \caption{Cellphones}   
    \label{fig:subfig_cellphones}
\end{subfigure}
\begin{subfigure}[b]{0.33\textwidth}
    \centering
    \resizebox{0.6\linewidth}{!}
    {
        \begin{tikzpicture}
        \pgfplotsset{
            scale only axis,
            label style={font=\normalsize}
        }
            \begin{axis}[
                xlabel=Training Epochs,
                xmin=0, xmax=100,
                xtick={20, 40, 60, 80},
                xticklabels={20, 40, 60, 80},
                ymajorgrids=true,
                ymin=0, ymax=0.8,
                legend style={
                }
            ]
            
            \addplot[smooth,color=blue,mark=triangle]
                coordinates {
                    (0,0.0435)
                    (20,0.3233)
                    (40,0.3298)
                    (60,0.3338)
                    (80,0.3315)
                };
            \addlegendentry{NDCG@10}
            
            \addplot[smooth,mark=square,red] plot coordinates {
                (0,0.0928)
                (20,0.5278)
                (40,0.5164)
                (60,0.5145)
                (80,0.5007)
            };
            \addlegendentry{HR@10}
            
            \end{axis}
        \end{tikzpicture}
    }
    \caption{Grocery}   
    \label{fig:subfig_grocery}
\end{subfigure}
\vspace{-15pt}
\caption{An illustration of the ranking performance changing with the training epochs.}
\label{fig:efficiency}
\vspace{-10pt}
\end{figure*}

\subsection{Global vs. Adaptive (RQ2)}

We claimed that it is important to design an adaptive NAS algorithm to realize architecture-level personalization for recommendation tasks. To verify the importance of this adaptive feature, we use a non-adaptive NAS, called NANAS. This is to replicate a regular NAS algorithm that learns a global architecture for all inputs on a specific task. The difference between MANAS and NANAS lies in the controller part. In NANAS, the predicate embeddings of the raw input variables do not involve in the search process. Instead, all the input variables $\textbf{e}_i$ are replaced with position embedding $\textbf{p}_i$, where $\textbf{p}_i$ is the representation of the $i$-th position in the logic expression. For example, suppose we are given two input sequences $s_1=\{v_1, v_2, v_3\}$ and $s_2=\{v_4, v_5, v_6\}$, MANAS search space includes the predicate embedding of the input variables $\{\textbf{e}_1, \textbf{e}_2, \textbf{e}_3, \textbf{e}_4, \textbf{e}_5, \textbf{e}_6\}$ as well as logical modules AND, OR, NOT. However, in NANAS, the search space only contains the position embedding $\{\textbf{p}_1, \textbf{p}_2, \textbf{p}_3\}$ and three logical modules. Since the search space does not relate to the input variables, the searched architecture is a global architecture that fit all the data of the given task. Here we have two observations from the NANAS results in Table~\ref{tb:results}:

(1) By comparing NANAS with NCR, we see that NANAS can consistently outperform NCR on all the tasks. Such results tell us that a searched neural logic architecture is better than a human-designed architecture. 
    
(2) By comparing NANAS with MANAS, we observe that MANAS compete NANAS on all the datasets over all the metrics. It empirically shows that an adaptive neural architecture is important for improving personalized recommendation quality.

\begin{figure}[t!]
    \begin{subfigure}[b]{0.32\linewidth}
        \centering
        \resizebox{\linewidth}{!}{
    \includegraphics[scale=0.3]{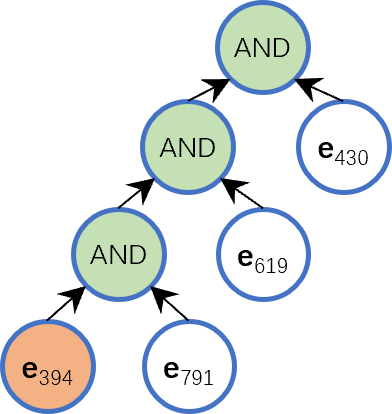}
        }
        \label{fig:sample_a}
    \end{subfigure}
    \begin{subfigure}[b]{0.32\linewidth}
        \centering
        \resizebox{\linewidth}{!}{
    \includegraphics[scale=0.45]{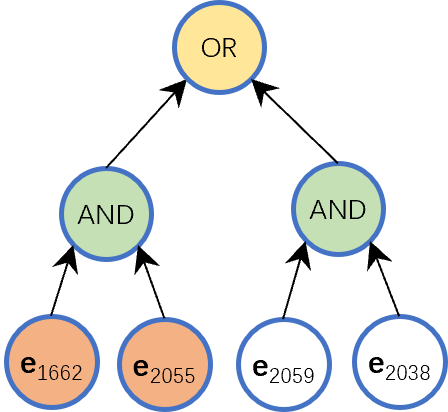}
        }
        \label{fig:sample_b}
    \end{subfigure}
    \begin{subfigure}[b]{0.32\linewidth}
    \centering
        \resizebox{\linewidth}{!}{
        \includegraphics[scale=0.4]{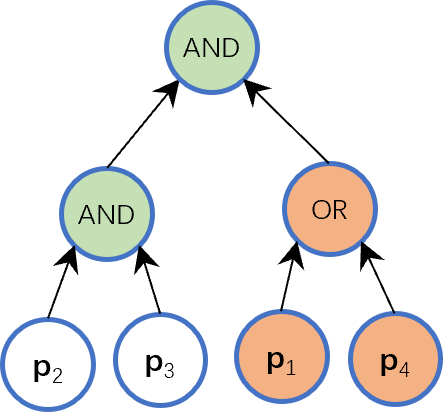}
        }
        \label{fig:sample_global}
    \end{subfigure}
\vspace{-10pt}
\caption{Examples of generated architectures on Beauty dataset. MANAS searched architectures (left) for the input sequence $\{\textbf{e}_{384}, \textbf{e}_{791}, \textbf{e}_{619}, \textbf{e}_{430}\}$ and (middle) for the input sequence $\{\textbf{e}_{1662}, \textbf{e}_{2055}, \textbf{e}_{2059}, \textbf{e}_{2038}\}$; NANAS generated the global architecture (right). The red-colored module means that its output should be negated as the input to its successor module.}
\label{fig:samples}
\vspace{-5ex}
\end{figure}

We illustrate the model generated architectures in Fig.(\ref{fig:samples}). It shows that our model can generate diverse architectures to adapt to different inputs. It helps neural logic reasoning to generalize to different inputs and logical expressions without the pain of manually designing suitable logical architectures.

\subsection{Importance of Exploration (RQ3)}
Exploration is one of the most important concepts in reinforcement learning. Without sufficient exploration, the learning process may not able to converge to a good solution. To verify the importance of exploration, we test a non-sampling version of MANAS. 

As mentioned in the previous section, the prediction of the next module is based on the probabilities in the logits vector, i.e., we treat all the probabilities as a distribution $\pi_\theta$ and the next module is sampled according to this distribution. The probability of each module being sampled at the current step is its corresponding probability value in the logits vector. This allows the searcher to explore potential cases that are not the best in the current step but may bring large rewards in the future, so as to avoid being trapped into a local optimal. To test the importance of exploration based on such sampling strategy, we test a non-sampling version of MANAS by doing the greedy selection, which always chooses the module with the max current probability as the next predicted module.
From the results in Table~\ref{tb:non_sample}, we find that MANAS has a significant improvement against the non-sampling model. This result shows the importance of doing exploration in the search process.

Since the exploration brings uncertainty in the architecture generation process, we sample 20 different architectures from the trained model for each input sequence and evaluate their performance. The results in Table~\ref{tb:non_sample} are the averaged results over the 20 samples. The small standard deviation reported in the table shows that our model can provide relatively consistent performance.

\subsection{Efficiency Analysis (RQ4)}
NAS training is time consuming especially when it comes to the adaptive neural architecture search. It is challenging to do batch training when the network architectures are different. To tackle this problem, we engineer the training of MANAS into two steps. Take the child network training process as an example. We first send the training data $\mathcal{D}_{train}$ into the controller to generate architectures for all the samples $s\in\mathcal{D}_{train}$. Since we limit the length of history to be exactly $n$ for each user-item interaction, we can guarantee that each sequence contains $n$ raw input variables. We assign a unique position index to each of the variables for each sequence. During the architecture generation stage, the sampler samples the position index for each input sequence and we can then use these position indices to assemble logical networks. For example, if we have two sequences $s_1 =\{\textbf{e}_1, \textbf{e}_2, \textbf{e}_3, \textbf{e}_4\}$ and $s_2=\{\textbf{e}_5, \textbf{e}_6, \textbf{e}_7, \textbf{e}_8\}$, we assign position index $0$ to $\textbf{e}_1$ and $\textbf{e}_5$, $1$ to $\textbf{e}_2$ and $\textbf{e}_6$, etc. When the controller generates the same sequence for both $s_1, s_2$ with $\{\textnormal{AND},0,1; \textnormal{OR},2,4; \textnormal{AND},3,5\}$, we can use the position index to locate the variable to create each layer of the logical network. Take the first sequence $s_1$ as an example, the corresponding logical network is $((\textbf{e}_1 \wedge \textbf{e}_2) \vee \textbf{e}_3) \wedge \textbf{e}_4$.

After obtaining all the architectures, we can group the sequences based on their sampled architectures. We can do batch training on those data that have the same architecture. We set the batch size to 256 for child network training. This number is an upper bound for the batch size. It is possible to have some architecture pattern groups whose number of sequences is less than the batch size. In that case, the minimum size of batches depends on the smallest group size of architecture patterns. For the controller training process, we apply the same approach to enable batch training. By setting the architecture generation batch to 1024 and the child network training batch to 256, we can achieve at least 5 times speed boost than non-batch training. To show the effectiveness of our batch training design, we report the change of recommendation performance in terms of training epochs in Fig.(\ref{fig:efficiency}). The results are reported based on the validation set and the sequence length is set to 4. We see that our model converges to a relatively optimal solution in around 30 epochs on all three datasets. To speed up the training process, we allow batch training for both controller and the searched child network. For each training epoch, we train the child network using only one whole pass of the training data and then train the controller network for 50 steps. Under these settings, one training epoch costs about 1.2 hours. 

During the inference stage, our model can generate the recommendation results efficiently. Specifically, the average inference time is $0.995\pm 0.054$ millisecond(ms), with $0.43\pm 0.14$ms for sampling the architecture and $0.57\pm 0.06$ms for generating the prediction. For reference, the inference time of SASRec is $0.67\pm 0.04$ms and BERT4Rec is $0.70\pm 0.02$ms. All the reported inference running time are based on a single record prediction.

\subsection{Influence of Sequence Length (RQ5)}
\label{sec:sequence_length}
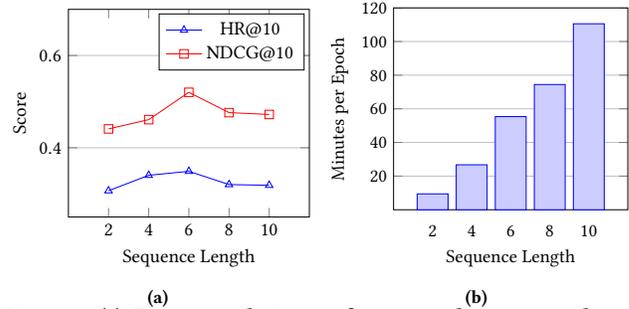
\begin{figure}
\begin{subfigure}[b]{0.49\linewidth}
        \resizebox{1\linewidth}{!}{
        \begin{tikzpicture}
\pgfplotsset{
    scale only axis,
    label style={font=\normalsize}
}

\begin{axis}[
  ymin=0.25, ymax=0.70,
  enlarge x limits=0.25,
  xlabel={Sequence Length},
  ylabel={Score},
  xtick={2, 4, 6, 8, 10},
  xticklabels={2, 4, 6, 8, 10},
  ymajorgrids=true,
]
\addplot[mark=triangle, blue]
  coordinates{
    (2,0.3069)
    (4,0.3404)
    (6,0.3490)
    (8,0.3203)
    (10,0.3187)
}; 
\addplot[mark=square, red]
  coordinates{
    (2,0.4409)
    (4,0.4610)
    (6,0.5202)
    (8,0.4762)
    (10,0.4722)
}; 

\legend{HR$@10$, NDCG$@10$}

\end{axis}
\end{tikzpicture}
}
\caption{}
        \label{fig:subfig_hist_len_score}
    \end{subfigure}
\begin{subfigure}[b]{0.49\linewidth}
    \centering
        \resizebox{1\linewidth}{!}{
            \begin{tikzpicture}
\pgfplotsset{
    scale only axis,
    label style={font=\normalsize}
}

\begin{axis}[
ybar,
ymin=0, ymax=120,
bar width=15pt,
enlarge x limits=0.25,
symbolic x coords={2, 4, 6, 8, 10},
  xlabel={Sequence Length},
  ylabel={Minutes per Epoch},
  xtick=data,
  xtick style={draw=none},
  ytick={20, 40, 60, 80, 100, 120},
  ymajorgrids=true,
  legend style={font=\fontsize{7}{5}\selectfont}
]
\addplot[draw=blue,fill=blue!20!white] 
coordinates{
(2, 9.45)(4, 26.72)(6, 55.37)(8, 74.45)(10, 110.55)}; 
\end{axis}

\end{tikzpicture}
}
        \caption{}   
        \label{fig:subfig_histlen_time}
\end{subfigure}
\vspace{-10pt}
\caption{(a) Recommendation performance change w.r.t the sequence length; (b) The histograms represent the architecture search time costs w.r.t different input sequence lengths. The time is given in minutes per epoch.
}
\label{fig:hist_len}
\vspace{-10pt}
\end{figure}

The length of input sequence could affect not only the recommendation performance but also the training time. In this subsection, we discuss how the recommendation performance and training time change with respect to the sequence length. We conduct 5 experiments by choosing the input sequence length from $\{2, 4, 6, 8, 10\}$. For example, we limit each training sample to have exactly 4 historical interactions when we refer to the sequence length as 4. One problem is that we cannot guarantee the five length settings will give us the same set of users. For example, a user with exactly 4 histories will not appear in the 6-history setting. In that case, we may have different training data for different sequence length experiments. To guarantee that all the comparisons are reasonable by training and evaluating models on the same data, we first filter the original dataset by only keeping those interactions that have at least 10 histories. Then we cut off the history sequence of each interaction for different sequence length experiments. In this case, we can guarantee that the performance as well as the training time differences only in the sequence length. Since the dataset used in this experiment is a filtered one, the training time and the performance reported in this subsection may be slightly different from previous results. The results are reported in Fig.(\ref{fig:hist_len}). 

From Fig.(\ref{fig:subfig_hist_len_score}), we have two observations. First, we find that the model can get recommendation performance gain by reasonably increasing the sequence length. This is because a longer sequence can potentially bring more information to help model capture user preferences. However, our second observation is that we cannot continue to get benefits from increasing the history length. Contrarily, the performance could be harmed when the sequence is too long. One reason is that a longer sequence could introduce noisy information to the model because some very early user histories may have very limited contribution or even be irrelevant to the current recommendation. In this case, a longer sequence does not help to improve the recommendation quality. On the other hand, the noise in the sequence could even result in the performance loss.

In Fig.(\ref{fig:subfig_histlen_time}), we plot the training time per epoch with respect to different sequence lengths. It shows that the training time is closely related with the sequence length. This is because a longer sequence represents a larger search space. By observing Fig.(\ref{fig:subfig_hist_len_score}) and (\ref{fig:subfig_histlen_time}) together, we can see that we need to carefully set the history length for MANAS. The limited information from the short sequence may prevent the model from capturing user preferences, thus resulting in low quality recommendations. However, longer sequence does not always help to improve the model performance. After reaching a certain threshold, longer sequences may have very limited or even no contribution to improve the performance but can greatly increase the training time.


\section{Conclusions and Future Work}
\label{sec:conclusion}
In this work, we propose to learn basic skills as neural modules and automatically assemble them into different models for solving a compositional number of different problems. Technically, we use intelligent recommender system as an example to demonstrate the idea and propose a Modularized Adaptive Neural Architecture Search (MANAS) framework, which automatically assembles logical operation modules into a reasoning network that is adaptive to the user's input sequence, and thus advances personalized recommendation from  the learning of personalized representations to the learning of personalized architectures for users. We enable neural-symbolic reasoning to generate flexible logical architectures, which make logical models adaptive to the diverse inputs without using human-crafted logical structure. The experimental results show that our design can provide significantly better prediction accuracy.
Besides, we also conduct experiments to show the importance of exploration in the architecture search process and the importance of learning adaptive architectures for the prediction and recommendation tasks. 

In this work, we considered three neural modules to demonstrate the idea of modularized adaptive neural architecture search. However, the proposed framework is general and can incorporate more neural modules such as neural predicates or other basic vision, language or recommendation modules, which we will explore in the future. Furthermore, in addition to the performance improvement, the problem-adaptive architecture generated by our framework may improve the model transparency, provide model explainability \cite{zhang2014explicit,zhang2020explainable,li2022from}, and enable automatic learning to define problems \cite{zhang2021problem}, which are important and promising to explore in the future.

\section*{Acknowledgments}
This work was supported in part by NSF IIS-1910154, IIS-2007907, IIS-2046457 and CCF-2124155. Any opinions, findings, conclusions or recommendations expressed in this material are those of the authors and do not necessarily reflect those of the sponsors.







\bibliographystyle{ACM-Reference-Format}
\balance
\bibliography{reference}

\appendix

\end{document}